\documentclass{article} 
\usepackage{preprint}
\usepackage[twoside,letterpaper,includeheadfoot,%
        layoutsize={8.125in,10.875in},%
                layouthoffset=0.1875in,%
                layoutvoffset=0.0625in,%
                left=1.25in,%
                right=1.25in,%
                top=45pt,%
                bottom=45pt,%
                headheight=0pt,
                headsep=10pt,%
                footskip=25pt]{geometry}

\usepackage{amsmath,amsfonts,bm}

\def\eqref#1{equation~\ref{#1}}

\def\1{\bm{1}}

\DeclareMathAlphabet{\mathsfit}{\encodingdefault}{\sfdefault}{m}{sl}
\SetMathAlphabet{\mathsfit}{bold}{\encodingdefault}{\sfdefault}{bx}{n}

\usepackage{hyperref}
\usepackage{url}
\usepackage{xcolor}
\usepackage{booktabs}
\usepackage{float}
\usepackage{adjustbox}
\usepackage{multirow}
\usepackage{pdflscape}
\usepackage{natbib}                
\usepackage{xargs}
\usepackage{ulem}

\usepackage{enumitem}
\setlist[itemize]{leftmargin=*}

\title{\LARGE \textbf{On the robustness of modeling grounded word learning through a child's egocentric input} }
\date{}
\author{ \large \textbf{Wai Keen Vong}$^1$ and \textbf{Brenden M. Lake}$^{1,2}$\\
\normalsize $^1$Center for Data Science, New York University\\
\normalsize $^2$Department of Psychology, New York University
}

\begin{document}
\maketitle
\begin{abstract}

What insights can machine learning bring to understanding human language acquisition? Large language and multimodal models have achieved remarkable capabilities, but their reliance on massive training datasets creates a fundamental mismatch with children, who succeed in acquiring language from comparatively limited input. To help bridge this gap, researchers have increasingly trained neural networks using data similar in quantity and quality to children's input. Taking this approach to the limit, \citet{vong2024grounded} showed that a multimodal neural network trained on 61 hours of visual and linguistic input extracted from just one child's developmental experience could acquire word-referent mappings. However, whether this approach's success reflects the idiosyncrasies of a single child's experience, or whether it would show consistent and robust learning patterns across multiple children's experiences was not explored. In this article, we applied automated speech transcription methods to the entirety of the SAYCam dataset, consisting of over 500 hours of video data spread across all three children. Using these automated transcriptions, we generated multi-modal vision-and-language datasets for both training and evaluation, and explored a range of neural network configurations to examine the robustness of simulated word learning. Our findings demonstrate that networks trained on automatically transcribed data from each child can acquire word-referent mappings, generalizing across videos, children, and image domains. These results validate the robustness of multimodal neural networks for grounded word learning, while highlighting the individual differences that emerge in how models learn when trained on each child's developmental experiences.
  
\end{abstract}

\section{Introduction}

Recent progress in large language models and multi-modal AI systems have led to fluent language generation and accurate object recognition, among other human-like abilities. These developments have naturally piqued the interest of many cognitive scientists and developmental psychologists, raising the question of whether these computational advances can help explain how children acquire language in the first few years of their lives. Yet, there is a striking asymmetry to reckon with: can machines trained to learn language from billions, even trillions, of word tokens provide insight into how children learn language from mere millions \citep{FrankTICS2023}? Addressing this disparity requires examining whether these language learning mechanisms can succeed under more realistic developmental constraints, combining these computational methods with richer and more ecologically valid datasets from developmental psychology.
 
Recent efforts towards this goal have examined what can be learned from linguistic (and other sensory) data that is comparable in quantity or kind to what children receive naturally \citep{huebner2021babyberta,wang2023finding,warstadt2024insights,qin2024systematic, vong2024grounded,hu2024findings,zhuang2024lexicon}. For example, \citet{huebner2021babyberta} examined language learning in neural networks from aggregated child-directed speech corpora \citep{MacWhinney1992TheCP}, demonstrating that models can acquire syntactic and semantic knowledge in the limited-data regime that is more representative of the kinds of input children receive. Similarly, the BabyLM challenge \citep{warstadt2024insights} examined what kinds of linguistic competencies can be learned from training language models on datasets consisting of 10M to 100M words \citep{hu2024findings,Charpentier2025BabyLMT3}. Separately, \citet{zhuang2024lexicon} studied how visual information can improve language learning, especially when using developmentally plausible amounts of data. 

Additional recent work has aimed to increase the realism of the training data by using egocentric, head-mounted camera recordings from developing children, such as SAYCam \citep{sullivan2022saycam} and BabyView \citep{long2024babyview}. These datasets provide a unique glimpse into what children actually see and hear over the course of language learning, providing a testbed for what can be learned by training models from this kind of naturalistic, high-quality input. Researchers have shown that certain kinds of syntax and semantics, as well as word-referent mappings, are all learnable, even under the extreme constraint of only using data collected from a single child \citep{wang2023finding, qin2024systematic, vong2024grounded, long2024babyview}.

In recent work most related to the current article, \citet{vong2024grounded} demonstrated that a multimodal neural network trained on vision-and-language data extracted from a single child could acquire word-referent mappings. Specifically, they used a subset of a single child in SAYCam \citep{sullivan2022saycam}, curating a dataset of paired child-directed utterances with corresponding video frames from 61 hours of video. This dataset was used to train a multimodal neural network utilizing two separate encoders for visual and linguistic inputs, combined with a contrastive objective to align embeddings from both modalities into a shared multimodal representation space. The model's knowledge of word-referent mappings was evaluated on two tasks involving 22 within-distribution categories and 64 out-of-distribution categories, achieving high performance on the former and modest performance on the latter. This work demonstrated that grounded word-referent mappings could be acquired from a subset of one child's experience from relatively generic neural networks. However, a major limitation is that its findings are reliant on the data from a single child (as only one child's data had transcripts at the time), raising questions about how robust or generalizable the findings are.

In this article, we examine the robustness of the findings reported in \cite{vong2024grounded} by utilizing the full 500 hours of video from the SAYCam dataset \citep{sullivan2022saycam} and training distinct models on each child's data. The main contributions of our study are:

\begin{itemize}
  \item We use automated speech recognition methods to fully annotate and transcribe approximately 500 hours of video from the SAYCam dataset, leading to multimodal datasets consisting of over 1.77M video frames paired with 266K utterances, a dataset 7 times larger than the previously available manually transcribed dataset.
  \item We assess the robustness of learning word-referent mappings from naturalistic visual-linguistic data by training separate vision-language models on each of the three different children in the SAYCam dataset from automated transcripts. Our results demonstrate how word-referent mappings are learnable across different datasets, despite varying degrees of visual-linguistic alignment from child to child.
  \item We examine how learned word-referent mappings generalize beyond training data, finding successful generalization across videos, across children, and across image domains, with some degree of transfer even under substantial domain shifts.
  \item We explore robustness across different language model configurations and training objectives, finding modest variations in performance, suggesting that dataset characteristics outweigh specific modeling or architectural choices.

\end{itemize}

\section{Datasets}

Our approach requires datasets that closely mirror the visual and linguistic experience from the perspective of developing infants. One dataset that fits this criterion is SAYCam \citep{sullivan2022saycam}, a video dataset capturing the egocentric, longitudinal perspective of 3 different children aged 6 to 32 months, collected via head-mounted cameras worn for approximately two hours per week. Parents were instructed to record naturalistically during everyday activities, resulting in footage covering mealtimes, play, reading, and outdoor activities. The three children (S - male, A - female, Y - male) were all from English-speaking families in the United States and Australia, each with a primary caregiver who was a psychologist. One child (S) was diagnosed with autism spectrum disorder at age 3, after the recording period; as of age 7, S was fully mainstreamed and did not require any special support. The other two children were typically developing.

While previous attempts to train language and multimodal models from SAYCam showed promise in answering questions around learnability \citep{wang2023finding, vong2024grounded, qin2024systematic}, their results were limited by the availability of manually transcribed natural language data, primarily to a subset of one of the three children's data. In this work, we expand the amount of available data for training by a factor of 7 through the use of automated speech transcription tools \citep{radford2023robust,bain2023whisperx}. This enabled the creation of new training and evaluation datasets for each of the three children in SAYCam, providing an opportunity to replicate our studies of simulated word learning with multiple children's data.

\subsection{Training Datasets}

\textbf{Transcribing videos}. In order to train separate vision-language models on each child's data, we needed a way to transcribe all of the speech from the raw videos. To efficiently transcribe the long-form audio from 500 hours of video from SAYCam, we used WhisperX \citep{bain2023whisperx}, a tool which leverages the transcription capabilities of OpenAI's Whisper model \citep{radford2023robust} (using the \texttt{v3-large} model with the language set to English), and applies further processing for efficient transcription and speaker diarization and word-level timestamps. This provided us with the transcribed text of each separate utterance, the start and end timestamps, and the predicted speaker label. Additional details surrounding the automatic transcription procedure and evaluation for speech transcription accuracy can be found in \cite{zhang2025metadata}.

Due to the inherent challenges of recording audio from a child's head-mounted camera, the speech intelligibility can vary greatly across contexts, especially for child-directed speech. Nevertheless, an analysis comparing a subset of manually vs. automatically transcribed speech content showed that it was relatively accurate for adult speech and also sufficiently similar to the previously manually transcribed subset \citep{zhang2025metadata}. Furthermore, although the relative timings for the detected start and end timestamps also showed some discrepancies, this is not necessarily a major issue: \cite{vong2024grounded} observed that imperfect timing information (in the manual transcriptions) were still sufficient for models to acquire word-referent mappings. 

Whisper was not a panacea for annotating the data: it showed relatively poor performance on diarization accuracy (correctly distinguishing between different speakers in a video). Therefore, we did not attempt to distinguish between different speakers (e.g. child-directed versus child-produced speech), and retained all of the transcribed utterances. Another limitation was from hallucinations in the Whisper transcriptions, which resulted in instances of repetitive utterances,\footnote{This included filtering repetitive phrases within a single utterance, as well as filtering adjacent utterances with the same content. In both cases, we only kept a single instance of the repeated utterance.}, requiring them to be filtered out. Additional pre-processing steps included removing all punctuation and lower-casing the transcribed text. Tokenization was performed at the word level using spaCy \citep{spacy2} and was subsequently used to create separate tokenizers from the transcribed vocabulary of each dataset split (after excluding any words with a frequency of two or fewer). After filtering and pre-processing, this resulted in 266K automatically transcribed utterances across the entire SAYCam dataset.

\textbf{Training datasets}. To create vision-language datasets for training, we used this set of transcribed utterances to extract video frames in their original resolution that were temporally aligned with each utterance. For each detected utterance, we used the detected starting timestamp to extract up to 16 frames from the video at a rate of 3.75 fps, corresponding to 4.27s of video. Each set of extracted video frames was paired with their corresponding utterance, forming the basis of our vision-language dataset for training. The total amount of available training data from automating the transcription process is roughly 7 times larger than what was previously available for a single baby via manual transcription \citep{sullivan2022saycam, vong2024grounded}. We split the utterances into five different dataset splits, making sure that each split uses data from at most one child. The five different splits are:

\begin{itemize}
\item \textbf{S-Whisper-2022}: This split is based on the subset of the Whisper utterances from the same set of videos that were previously manually transcribed for baby S, as described in \cite{sullivan2022saycam}, and used in \cite{vong2024grounded}.
\item \textbf{S-Whisper-Disjoint}: This split uses the complement of videos from \textbf{S-Whisper-2022}, the set of transcribed utterances from new videos from baby S that were not previously manually transcribed.
\item \textbf{S-Whisper}: This split uses the full set of Whisper utterances from baby S's videos.
\item \textbf{A-Whisper}: This split uses the full set of Whisper utterances from baby A's videos.
\item \textbf{Y-Whisper}: This split uses the full set of Whisper utterances from baby Y's videos.
\end{itemize}

For each dataset split, videos were partitioned into independent train, validation, and test sets. Summary statistics for each of these splits are shown in Table~\ref{tab:dataset-descriptives}.

\subsection{Evaluation Datasets}
\label{sec:evaluation-datasets}

\begin{figure}[t]
  \centering
  \includegraphics[width=\textwidth]{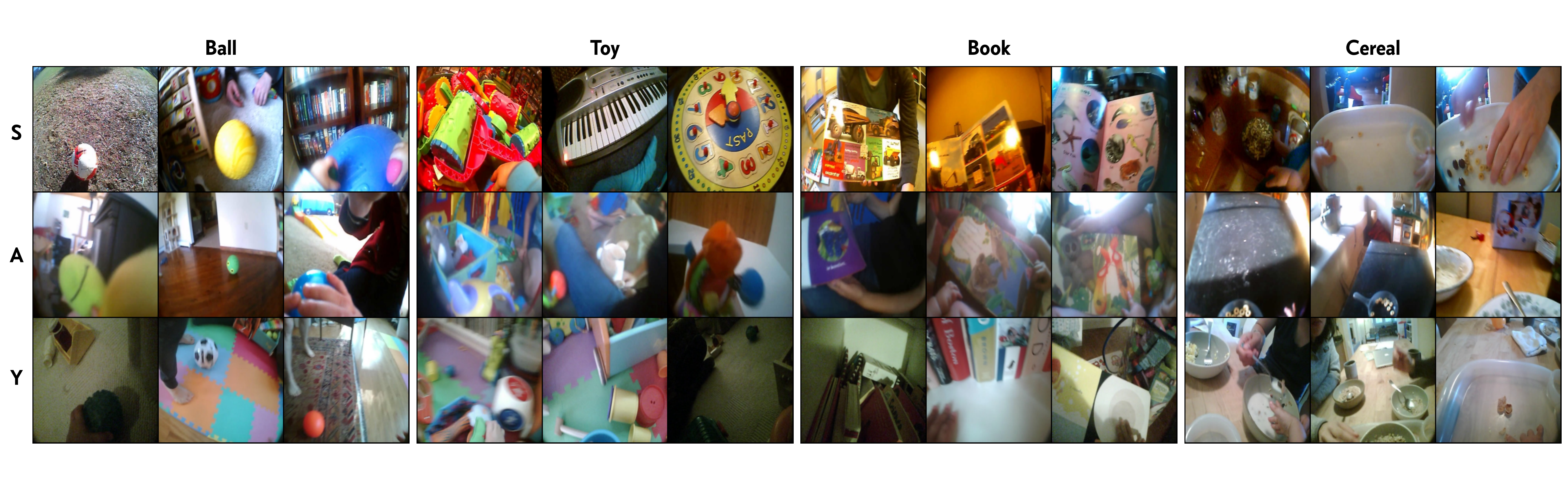}
  \caption{\textbf{Evaluation frames from Labeled-S-V2, Labeled-A and Labeled-Y.} Here we present three randomly selected evaluation frames from four different evaluation categories (ball, toy, book and cereal), with each row indicating frames derived from each of the three children in the SAYCam dataset.}
  \label{fig:evaluation-frames}
\end{figure}

For model evaluation, we primarily focus on image classification tasks, which are based on the looking-while-listening paradigm from early word learning studies \citep{golinkoff1992young}. In this evaluation procedure, models are presented with a target word along with four images (one belonging to the target category, and three distractors from other randomly sampled categories). The task is to select the image corresponding to the target word, where these labeled frames are sourced either from SAYCam directly, or other image datasets. Our model evaluations focus primarily on concrete nouns, as they are very frequent in training and straight-forward to construct evaluations for.\footnote{Other lexical categories, while potentially learnable, pose additional evaluation challenges in egocentric videos. For example, verbs often co-occur with parents narrating their child's actions, such that actions like ``run'' and ``jump'' are primarily associated with motion-blurred frames, while adjectives involving more abstract or relational properties are less reliably grounded directly in visual features.} We test our models on a number of different evaluation datasets:

\begin{itemize}
    \item \textbf{Labeled-S}: This first evaluation set consists of 22 different visual categories derived from baby S's egocentric video data, covering the most common kinds of objects and places both seen and spoken about \citep{orhan2020self, vong2024grounded}. For each category, there are 100 separate evaluation trials for a total of 2200 evaluation trials. The full set of categories can be found in Appendix~\ref{app:eval-categories}. This evaluation dataset is used to measure \textbf{within-child generalization}.
    \item \textbf{Labeled-S-V2}, \textbf{Labeled-A} and \textbf{Labeled-Y}: These three additional evaluation datasets were generated using frames from each child's own test split, enabling us to measure both \textbf{within-child} and \textbf{cross-child generalization}. \footnote{For Labeled-S-V2, due to some differences in dataset splitting, there was some partial overlap between the videos used for extracting frames for training and evaluation.} These evaluation datasets were constructed using a mixture of automatic and manual filtering (see Appendix~\ref{app:eval-datasets} for additional details), resulting in the same set of 41 object categories for each child's evaluation dataset.\footnote{The 41 categories partially overlap with, but are not a superset of, the original 22 Labeled-S categories. These 41 categories were selected based on common visual concepts present to all three children's environments.} Each evaluation dataset consists of 100 trials per category, for a total of 4100 evaluation trials. These 41 categories cover common visual concepts encountered by each child in SAYCam, including body parts (arm, hair, knee), clothing (shirt, jeans, shoe), living spaces (kitchen, porch, table), objects (ball, bottle, boy) and food (banana, cereal). A limited set of evaluation frames are shown in Figure~\ref{fig:evaluation-frames}, and the full set of categories can be found in Appendix~\ref{app:eval-categories}.
    \item \textbf{Konkle Objects}: This evaluation dataset consists of 60 visual categories across 1,784 trials derived from naturalistic photographs of everyday objects on white backgrounds \citep{konkle2010conceptual}, enabling us to measure \textbf{out-of-distribution generalization}. Similar to the above evaluation datasets, these 60 categories are also present in each of the children's vocabularies, facilitating cross-dataset comparisons.
\end{itemize}

\section{Models}

\begin{figure}[t]
  \centering
  \includegraphics[width=0.95\textwidth]{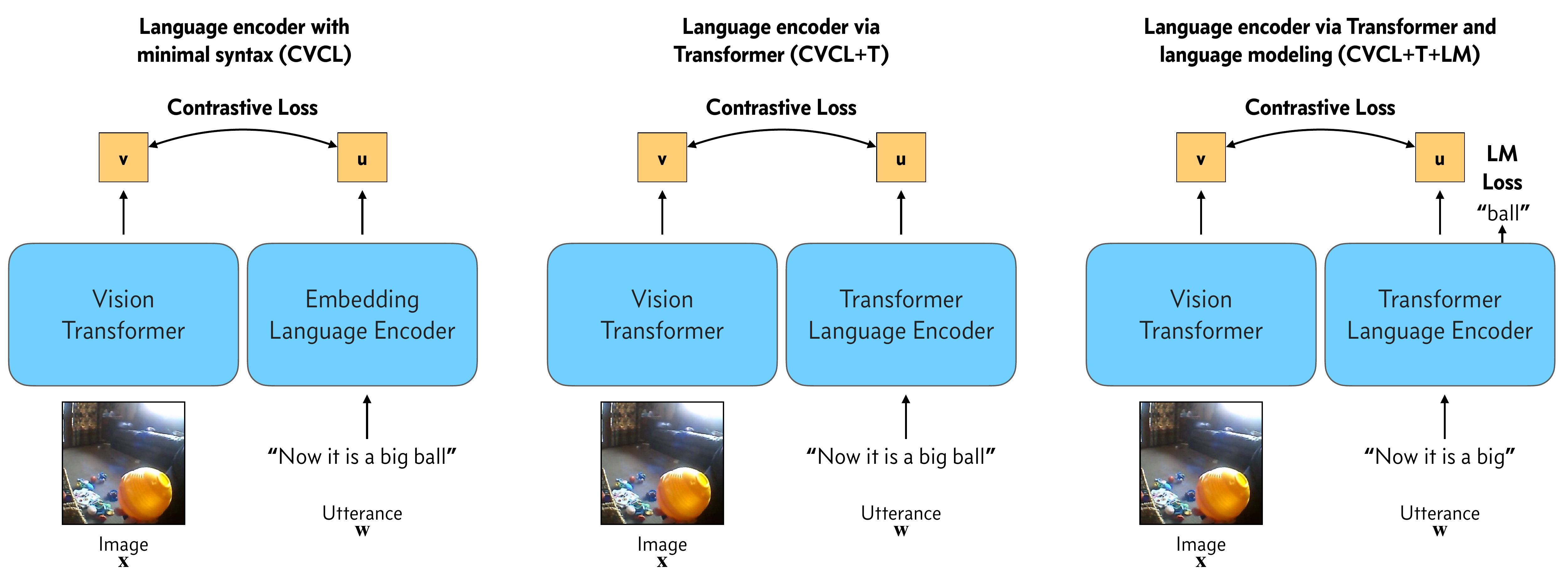}
  \caption{\textbf{Model architectures.} We explore three different multi-modal architectures. The first two architectures utilize a contrastive loss with either a simple Embedding layer with averaging across words (CVCL), or a 2-layer Transformer Decoder (CVCL+T). The third is also a 2-layer Transformer Decoder, but incorporates an additional language modeling loss head (CVCL+T+LM), with an additional weight parameter to balance the two losses. All models use a pre-trained Vision Transformer as their vision encoder, trained only from the visual data from each child, along with a learned, absolute positional embedding scheme in the language encoder.}
  \label{fig:model-architecture}
\end{figure}

We begin by describing the CVCL (Child's View for Contrastive Learning) model from \cite{vong2024grounded} before describing the variants we explore in this work. As described in the previous section, the input provided to the model is in the form of pairs of video frames and child-directed utterances. From this input, we need a method to align the representations from image frames with the representations of corresponding utterances, thereby linking visual referents to their corresponding words. To accomplish this feat, CVCL trains two neural networks jointly, learning to match image frames with their corresponding utterances. One of these networks (the image encoder) embeds video frames, while the other network (the language encoder), separately embeds utterances, projecting both into a shared multimodal representation space. This operationalizes the cross-situational learning approach from cognitive science \cite{yu2007rapid} through the lens of modern neural networks.

Specifically, CVCL's vision encoder is designed to take in static video frames, passing them through a ResNeXt CNN architecture \citep{xie2016resnext} that was pre-trained solely from the visual input from this child, and frozen except for a learnable linear projection head, to obtain a single embedding for each frame. CVCL's language encoder is designed to take in transcribed child-directed utterances, passing each word through a single Embedding layer separately and then averaging across each word's embedding in the utterance, again to obtain a single embedding for each utterance. Both encoders are jointly trained by passing embeddings into a contrastive loss, bringing closer matched embeddings from paired video frames and utterances, while separating mismatched embeddings from different video frames and utterances, similar to CLIP \citep{radford2021learning}. This process of joint associative and representation learning produces a shared multimodal representation space, where neural similarity between words and referents determines the strength of their mapping.

We systematically explore three different architectural modifications to CVCL to examine their effects on grounded word learning. Each of these variants utilizes the same underlying vision encoder architecture, but varies in the choice of language encoder and learning objective, each of which is illustrated in Figure~\ref{fig:model-architecture}.

\subsection{Vision Encoder}
For embedding static video frames, we replaced the ResNeXt-CNN in CVCL with a vision transformer (ViT-B/14) \citep{dosovitskiy2020image}, which is separately pre-trained only using visual data from the SAYCam dataset. This pre-training step is performed separately for each child's visual data, using the self-supervised approach from DINO \citep{orhan2024learning,caron2021emerging}. From the set of frames associated with a given utterance (up to 16 sampled frames), we always sample one frame randomly to be paired with the corresponding utterance. During training, data augmentation is performed by performing a random crop to each video frame, followed by resizing the frame to $224 \times 224$ pixels, and occasionally horizontally flipping the frame. During evaluation, frames are instead always resized to 256 pixels along the minor edge, followed by a $224 \times 224$ centered crop.

We obtain the output of the vision transformer from the \texttt{[CLS]} token, a single embedding of size 768. This embedding is fed through a learnable linear projection head, followed by a layer norm and a dropout layer, such that the vision encoder returns a single embedding of size $D=512$. Similar to \cite{vong2024grounded}, the bulk of the vision encoder is frozen during training, except for the learnable linear projection head.

\subsection{Variant 1: Language encoder with minimal syntax (CVCL)}
In this model variant, we incorporate word-level positional embeddings into the language encoder to provide some sensitivity to word order, laying the groundwork for additional transformer-based variants described below. Because this is a relatively minor change from the original CVCL model, we still refer to this variant as CVCL throughout. \\

\noindent \textbf{Language Encoder (Embedding)}: We used the same Embedding encoder from \cite{vong2024grounded}, which extracts separate embeddings for each word in an utterance, and computes the average across all word embeddings to produce a single utterance embedding of size $D = 512$. In this variant, we add a learned, absolute positional embedding to each token embedding before the averaging step.\footnote{Because these learned positional embeddings are averaged across all words in the utterance, utterances of the same length will have the same positional embeddings, regardless of word order. We include positional embeddings for consistency with the transformer-based variants, which preserve finer-grained word-level positional information through the attention mechanism.} Prior to passing utterances into the language encoder, tokenization was performed using a word-level tokenizer built via spaCy. \\

\begin{figure}[t]
  \centering
  \includegraphics[width=0.8\textwidth]{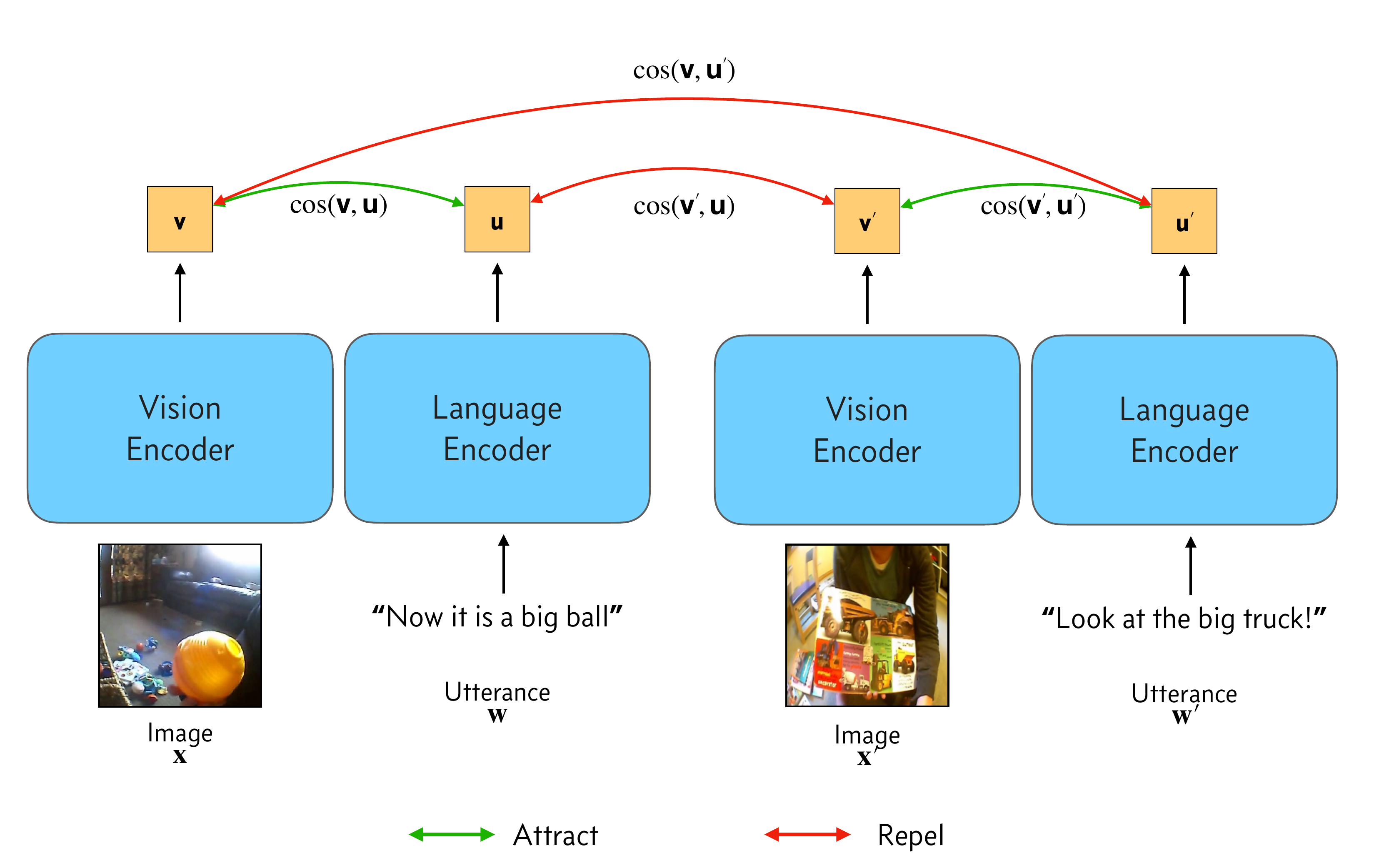}
  \caption{\textbf{Visualization of the contrastive objective.} Given multiple pairs of image frames and utterances, we obtain embeddings via their respective encoders. The contrastive objective aims to maximize the cosine similarity of these matched embeddings (shown in green), while also minimizing the cosine similarity of mismatched embeddings (shown in red).}
  \label{fig:contrastive-objective}
\end{figure}

\noindent \textbf{Learning Objective (Contrastive)}: Following a standard approach to training vision-language models \citep{radford2021learning, jia2021scaling}, we employ an InfoNCE contrastive learning objective, which aims to bring together frame embeddings with their corresponding utterance embeddings in a mini-batch (that is, encouraging a larger dot product between the vectors), while separating all other mismatching frame and utterance embeddings (smaller dot product between the vectors). First, we obtain visual embeddings for each frame, $\mathbf{v_i} = f_{\theta}(\mathbf{x_i})$ as well as corresponding text embeddings, $\mathbf{u_i} = f_{\phi}(\mathbf{w_i}).$ Then, the contrastive objective aims to match each frame with its corresponding utterance:
\begin{equation}
  L_{\text{frame}} = -\frac{1}{N} \sum_{i}^{N} \log \frac{\exp (\mathbf{v_i}^T \mathbf{u_i} / \tau)}{\sum_{j=1}^N \exp(\mathbf{v_i}^T \mathbf{u_j} / \tau)},
\end{equation}
as well as matching each utterance with its corresponding frame:
\begin{equation}
  L_{\text{utterance}} = -\frac{1}{N} \sum_{i}^{N} \log \frac{\exp (\mathbf{u_i}^T \mathbf{v_i} / \tau)}{\sum_{j=1}^N \exp(\mathbf{u_i}^T \mathbf{v_j} / \tau)}.
\end{equation}
These two losses are then summed together and averaged:
\begin{equation}
  L_\text{contrastive} = \frac{1}{2} L_{\text{frame}} + \frac{1}{2} L_{\text{utterance}}.
\end{equation}
The contrastive objective is visualized in Figure~\ref{fig:contrastive-objective}.

\subsection{Variant 2: Language encoder via Transformer (CVCL+T)}

In this variant, we examine the effect of using a more powerful language encoder. In the previous variant, the language encoder treats the importance of each word in an utterance equally, regardless of its context. To test this, we replace the Embedding-based language encoder with a 2-layer Transformer Decoder architecture instead, enabling it to create contextualized word representations by dynamically weighting contributions from surrounding words. We refer to this variant as \textbf{CVCL+T} (where +T indicates the use of the Transformer-based language encoder). \\

\noindent \textbf{Language Encoder (Transformer)}: We used a GPT-2 transformer decoder as our language encoder \citep{radford2019language}. Since GPT-2 uses a causal self-attention mask, the embedding from the last hidden state contains a contextualized aggregate representation of all of the preceding words in the utterance, allowing us to treat this representation as an utterance encoding that is further refined via contrastive learning with the vision encoder. All GPT-2 transformer models use 2 layers, 8 attention heads. We use the embedding of the last hidden state (corresponding to the \texttt{<eos>} token) as the output embedding, which is also of size $D = 512$. \\

\noindent \textbf{Learning Objective (Contrastive)}: This variant utilizes the same contrastive learning objective as Variant 1.

\subsection{Variant 3: Language encoder via Transformer and language modeling}

In this variant, we examine the effect of exposing the model to richer linguistic signals via an auxiliary language modeling objective. The original contrastive objective purely aims to maximize visual-text alignment, but may be missing out on some of the other statistical regularities present in the structure of language. To test this, we augment the contrastive objective from CVCL+T with an auxiliary language modeling objective, training the model to predict each next word in an utterance. We refer to this variant as \textbf{CVCL+T+LM} throughout the paper, where the +LM indicates the addition of the language modeling objective to the base Transformer architecture. \\

\noindent \textbf{Language Encoder (Transformer)}: This variant utilizes the same 2-layer Transformer architecture as Variant 2 for its language encoder. For the variant, we utilize the unembedding layer (which shares weights with the embedding layer) to generate the decoder's logits for predicting the next word in a given utterance. \\

\noindent \textbf{Learning Objective (Contrastive with Language Modeling)}: In addition to the contrastive learning objective, for this variant, we incorporate an additional language modeling loss, similar to \citep{zhuang2024lexicon}. The equation is given by:
\begin{equation}
  L_{\text{joint}} = L_{\text{LM}} + \lambda_c L_{\text{contrastive}}
\end{equation}
We scale the contrastive loss relative to the language modeling loss by $\lambda_c$, which controls the trade-off between vision-language alignment and language modeling. Following \cite{zhuang2024lexicon}, we set $\lambda_c = 0.3$ across all experiments with this variant, and validated this choice through a hyperparameter sweep. Since each child's training data is shuffled and presented over multiple epochs, $\lambda_c$ represents a consistent weighting of these two objectives, independent of the child's age in any given training example.

\subsection{Training Details}

Each model was trained up to 100 epochs using the AdamW optimizer, with a learning rate of 1e-4 or 1e-5 depending on the model type. The learning rate schedule involved 5000 steps of linear warmup, followed by cosine annealing until the end of training. Early stopping was performed using the validation loss.\footnote{To better track learning, we filter each validation set with CLIP to only contain frame-utterance pairs with high CLIP similarity scores. See Appendix~\ref{app:clip-validation-sets} for additional details.} For the vision encoder, we followed the specifications in \cite{vong2024grounded,orhan2024learning}: we used the augmentations as in the vision encoder pretraining (described above), we sampled a randomly associated frame with each utterance, and we froze the vision encoder during training (except for the projection head). For the language encoder, we use a maximum sequence length of 48. To train the model, we use a fixed temperature parameter in the contrastive loss with $\tau = 0.07$, with varying batch sizes depending on the model configuration. Vision dropout, language dropout and weight decay were all set to 0.1. We performed hyperparameter search over batch size, learning rate, vision dropout, language dropout, weight decay, and $\lambda_c$. For each dataset and model configuration, we trained 3 separate models with different random seeds. A full comparison of the various architecture and training details is shown in Table~\ref{tab:model-configurations}.

\section{Results}

One of the main evaluation methods for the original CVCL model was examining its zero-shot classification performance, specifically using the Labeled-S evaluation dataset, as described in Section~\ref{sec:evaluation-datasets}. This evaluation dataset consists of 22 different categories, with 100 4-way classification trials per category. Previously, \citet{vong2024grounded} showed that an Embedding-based multimodal neural network (CVCL) trained on the subset of manually transcribed data from baby S achieved 61.6\% classification accuracy on this dataset. In this section, we compare and contrast the performance of the original CVCL model against additional variants of CVCL trained using Whisper-derived transcriptions.

\subsection{Finding 1: Comparable performance for models trained with automatic transcriptions vs. manual transcriptions}

\begin{figure}
  \centering
  \includegraphics[width=0.9\textwidth]{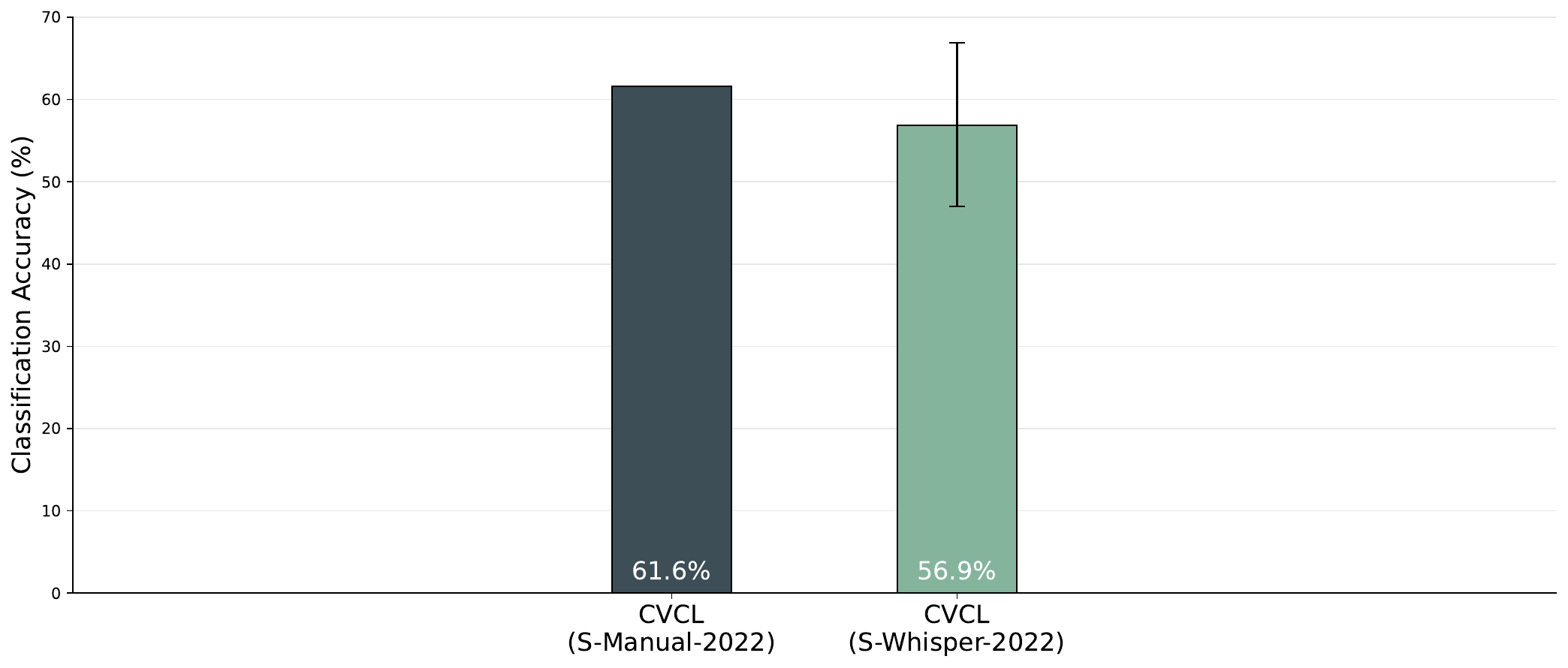}
  \caption{\textbf{Labeled-S Classification Performance between CVCL trained with manual transcriptions (S-Manual-2022) vs. trained with Whisper transcriptions (S-Whisper-2022).} Note that the version of CVCL in the right plot utilizes a pre-trained vision transformer (ViT) instead of the CNN (ResNeXt), as well as a positional encoding scheme in the language encoder. Models were trained with three random seeds. Error bars show bootstrapped 95\% confidence intervals over category-level accuracies.}
  \label{fig:S-Whisper-2022}
\end{figure}

We first evaluated whether automatic transcriptions from Whisper would be comparable to manual transcriptions for training multimodal neural networks. These models were trained using a subset of Whisper transcribed utterances belonging to the same subset of videos that were previously manually transcribed and used to train CVCL, enabling a direct comparison of Whisper's transcription quality. While initial analyses showed that the word frequency of the 22 categories from the Labeled-S evaluation showed a high correlation between the manually transcribed vs. Whisper transcriptions \citep{zhang2025metadata}, this did not provide any information about the quality of the Whisper-detected timestamps nor their temporal alignment with visual referents present in the dataset.

As can be seen in Figure~\ref{fig:S-Whisper-2022}, our results show that training CVCL on Whisper transcriptions provides comparable performance to CVCL, achieving a classification accuracy of 56.9\%, compared to 61.6\% from the original model. While there was a slight drop in performance compared to the original CVCL model, our results show that word-referent mappings are indeed learnable from Whisper-transcribed utterances, despite some of its limitations previously observed in transcription quality \citep{zhang2025metadata}. There are two additional reasons why the quality of Whisper transcriptions may have been sufficient to reach a similar level of performance as CVCL. First, rather than requiring perfect transcriptions for every word from every child-directed utterance, we only evaluate on a subset of the most prominent and commonly spoken nouns, for which there were a sufficient number of examples to learn from. Second, because of the relatively slow timescales at which visual information changes from the child's perspective, large differences between relevant visual referents and detected utterance timings may not have been as important. This robustness to small mis-alignments in timing was observed with CVCL due to how some of its utterances were pre-processed during training.

\subsection{Finding 2: Comparable performance for models trained on disjoint transcriptions}

\begin{figure}[t]
  \centering
  \includegraphics[width=0.9\textwidth]{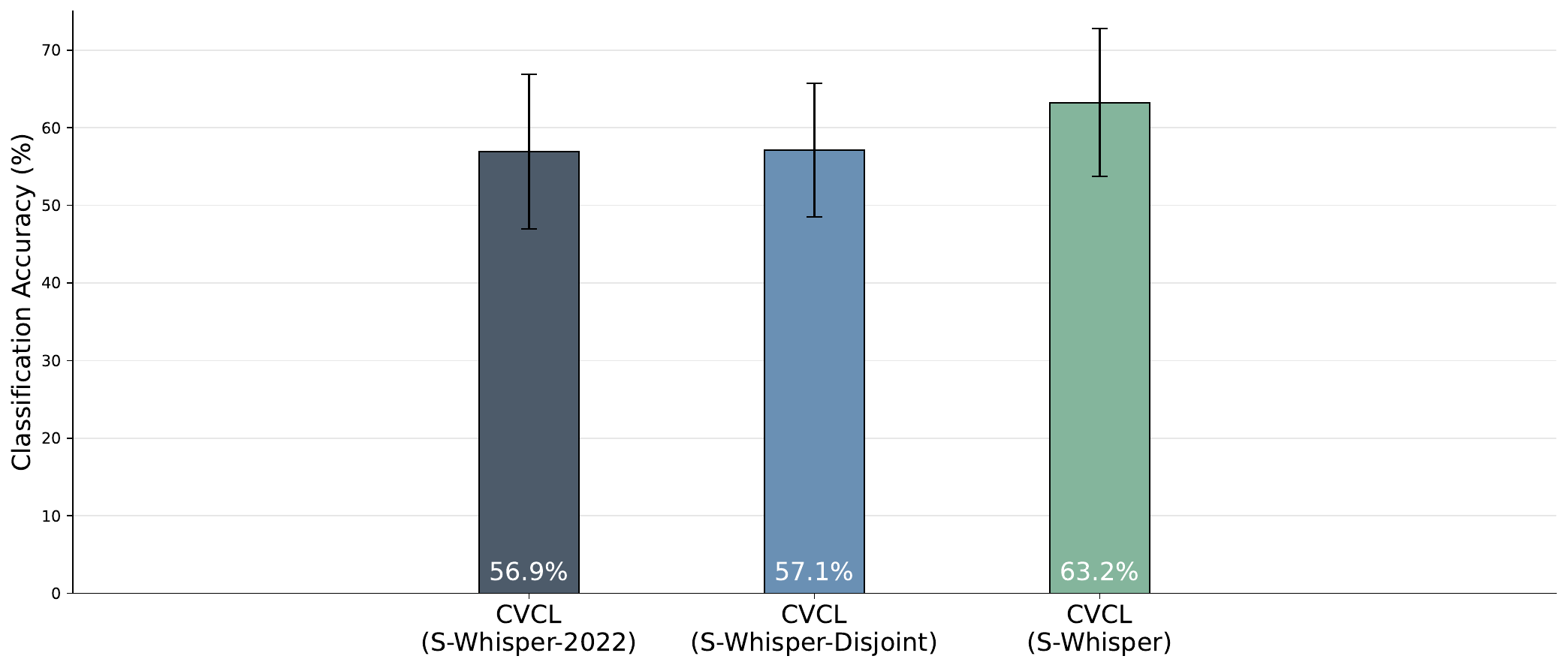}
  \caption{\textbf{Labeled-S Classification Performance between CVCL trained on different combinations of Whisper transcriptions.} The left-most bar (S-Whisper-2022, same as Figure~\ref{fig:S-Whisper-2022}), represents classification performance when trained on transcriptions from the set of previously manually transcribed videos. The middle bar (S-Whisper-Disjoint) represents performance when trained only using transcriptions from new and non-overlapping videos from baby S. Finally, the right-most bar (S-Whisper) represents performance when trained using all of the combined Whisper-transcribed data from original and new videos. Models were trained with three random seeds. Error bars show bootstrapped 95\% confidence intervals over category-level accuracies.}
  \label{fig:s-whisper-disjoint-and-combined}
\end{figure}

To examine whether our model genuinely learns generalizable word-referent mappings, we conducted a follow-up experiment where the frames for training versus evaluation are completely disjoint. This contrasts with the previous experiment, where training and evaluation frames were independently extracted, but from the same set of videos. We trained a distinct set of models using the \textbf{S-Whisper-Disjoint} split, which utilizes Whisper transcriptions derived from an independent and non-overlapping set of videos from baby S, distinct from the videos used in both the \textbf{S-Whisper-2022} videos and the Labeled-S evaluation set. Since all of the training data in this split comes from entirely new videos from baby S that were not previously manually transcribed, this enables us to examine whether the learned word-referent mappings generalize to the Labeled-S dataset. In this instance, the frames from Labeled-S are now considered as new, previously unseen visual instances of these categories that were not seen at all during training for the S-Whisper-Disjoint models, and can be thought of as a form of within-child generalization. This kind of exploration was not previously possible due to the limited available set of manually transcribed videos that were used for both training and evaluation (although the construction of each respective dataset was independent).

The results are shown in Figure~\ref{fig:s-whisper-disjoint-and-combined} (middle bar), and here we find that CVCL, when trained on non-overlapping data, achieves a comparable classification accuracy of 57.1\%, demonstrating robust within-child generalization to entirely new videos from the same child. Why do these models generalize well in this fashion? Despite being trained on a different set of videos, fundamentally, they are drawn from the same distribution of videos captured from a head-mounted camera in the same environment as the original manually transcribed videos for baby S, with similar objects and utterances across different videos. However, it is also the case that the contrastive objective forces multimodal representations to be robust and generalizable \citep{radford2019language,vong2024grounded}, guiding the models to learn the semantic relationship for different word-referent mappings, enabling broad generalization rather than overfitting or memorizing specific instances.

\subsection{Finding 3: Classification performance increases with additional training data}

To determine whether increased training data could improve model performance, we trained models that used the maximum amount of training data available from baby S. These models were trained on the \textbf{S-Whisper} split, which was constructed using a combination of the Whisper transcriptions from \textbf{S-Whisper-2022} and \textbf{S-Whisper-Disjoint}, resulting in approximately 3.5 times more data than used to train the original CVCL model (see Table~\ref{tab:dataset-descriptives} for additional details). 

As shown in Figure~\ref{fig:s-whisper-disjoint-and-combined} (right bar), we find that a CVCL model trained on this combined dataset split achieves an overall classification accuracy of 63.2\%. While this outperforms CVCL, the gains in accuracy are relatively minimal given the large relative increase in training data. There are several possible explanations for the limited benefit to more training data. One reason, similar to what was discussed in the previous section, is that the additional data from the newly transcribed videos in \textbf{S-Whisper-Disjoint} are being drawn from the same overall distribution of scenes and utterances, and therefore only providing a modest boost to the model's generalization capabilities. An alternative explanation is that some categories from the Labeled-S evaluation set are sufficiently noisy or challenging, not just for CVCL or the variants we explore here, but for all models. For instance, \cite{vong2024grounded} evaluated CLIP ViT-L/14 on this dataset and found its zero-shot classification performance was only marginally higher (66.7\%) despite being a strong vision-language foundation model, with CLIP also performing relatively poorly for a number of ambiguous categories like ``floor'' and ``hand''.

\subsection{Finding 4: Transformer-based language encoders achieve similar levels of classification performance}

\begin{figure}[t]
  \centering
  \includegraphics[width=0.95\textwidth]{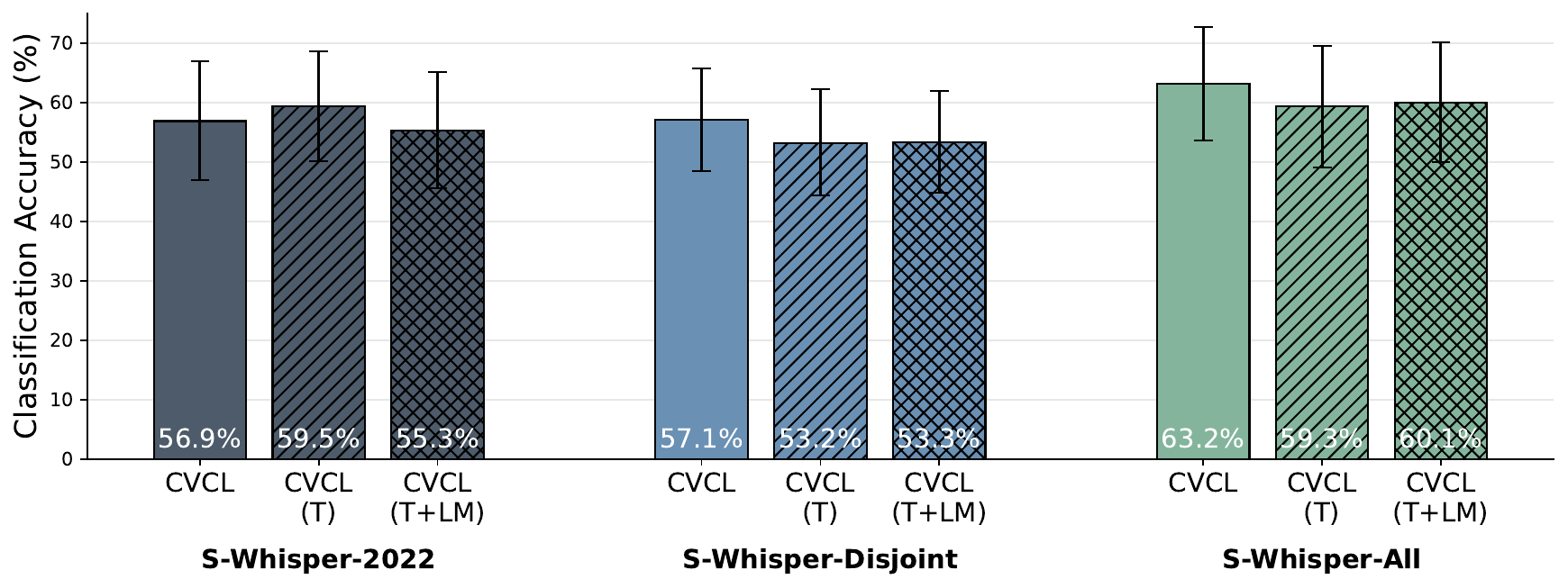}
  \caption{\textbf{Labeled-S Classification Performance between different combinations of Whisper transcriptions, across three different model configurations.} The left-most bar in each group represents performance from Figure~\ref{fig:s-whisper-disjoint-and-combined}. The two additional bars in each group explored modifications to the language encoder, either replacing the Embedding encoder with a Transformer (T) or with a Transformer and an additional language modeling head (T+LM). Models were trained with three random seeds. Error bars show bootstrapped 95\% confidence intervals over category-level accuracies.}
  \label{fig:transformers}
\end{figure}

To examine whether more sophisticated language encoders could improve grounded word learning, we replaced the Embedding-based language encoder with two variants involving a 2-layer Transformer. One of these variants used the same contrastive loss alone (CVCL+T), while the other variant combined this contrastive loss with an additional language modeling loss (CVCL+T+LM). These additional language encoder variants were chosen to examine whether other approaches for encoding utterances would be helpful for grounded word learning. The results in Figure~\ref{fig:transformers} show that the Transformer-based language encoders show a similar performance compared to CVCL across the different splits, although no particular variant is consistently better or worse. One reason we may not have seen any additional advantages is that our classification-based evaluation directly measures vision-language alignment, which all models are trained to optimize for via the contrastive objective. In a separate evaluation, we also examined word similarity across models (see Appendix~\ref{app:word-similarity-evaluation}), finding that the models trained via a joint contrastive and language modeling objective demonstrated stronger correlations with human similarity judgments than a contrastive objective alone.

\subsection{Finding 5: Robust evidence of learnability from training models across different children's datasets}

To test whether our findings generalize beyond a single child's linguistic environment, we examined model performance across multiple children's datasets. This addresses one of the key limitations from our previous analyses, which focused exclusively on learning from just one child's data. While it has been sufficient to demonstrate learnability from relatively generic learning mechanisms, it fails to address how generalizable and robust these findings are. Here, we leverage Whisper's ability to generate automatic transcriptions for all babies from the SAYCam dataset: S, A and Y. In this analysis, we train models on either \textbf{S-Whisper}, \textbf{A-Whisper}, or \textbf{Y-Whisper}.

In order to properly evaluate these models trained on other children's data, we also required a separate set of evaluation trials that were drawn from each child's videos respectively. This set of evaluations also consists of 4-way classification trials, but generated from a larger set of 41 categories, see Appendix~\ref{app:eval-categories} for additional dataset creation details. Notably, these evaluations were designed in a similar fashion to the \textbf{S-Whisper-Disjoint} split described above: the evaluation frames were generated from the ``test'' splits of each child's data, such that they were independent from anything seen during training or validation, enabling the measurement of within-child generalization.\footnote{However, due to differences in dataset splitting, the new set of evaluation frames derived from baby S are not completely independent from training on S-Whisper, but they are for the A-Whisper and Y-Whisper datasets.} These evaluation datasets all contain the same set of 41 categories that were present in all three children's visual and linguistic environments, enabling both a more comprehensive and comparable method of comparison for grounded word learning across children. Furthermore, we generate a different evaluation dataset from each child's test frames and only evaluate models trained on that child's data to their corresponding evaluation set, which we call Labeled-S-V2, Labeled-A and Labeled-Y respectively.

\begin{figure}[t]
  \centering
  \includegraphics[width=0.95\textwidth]{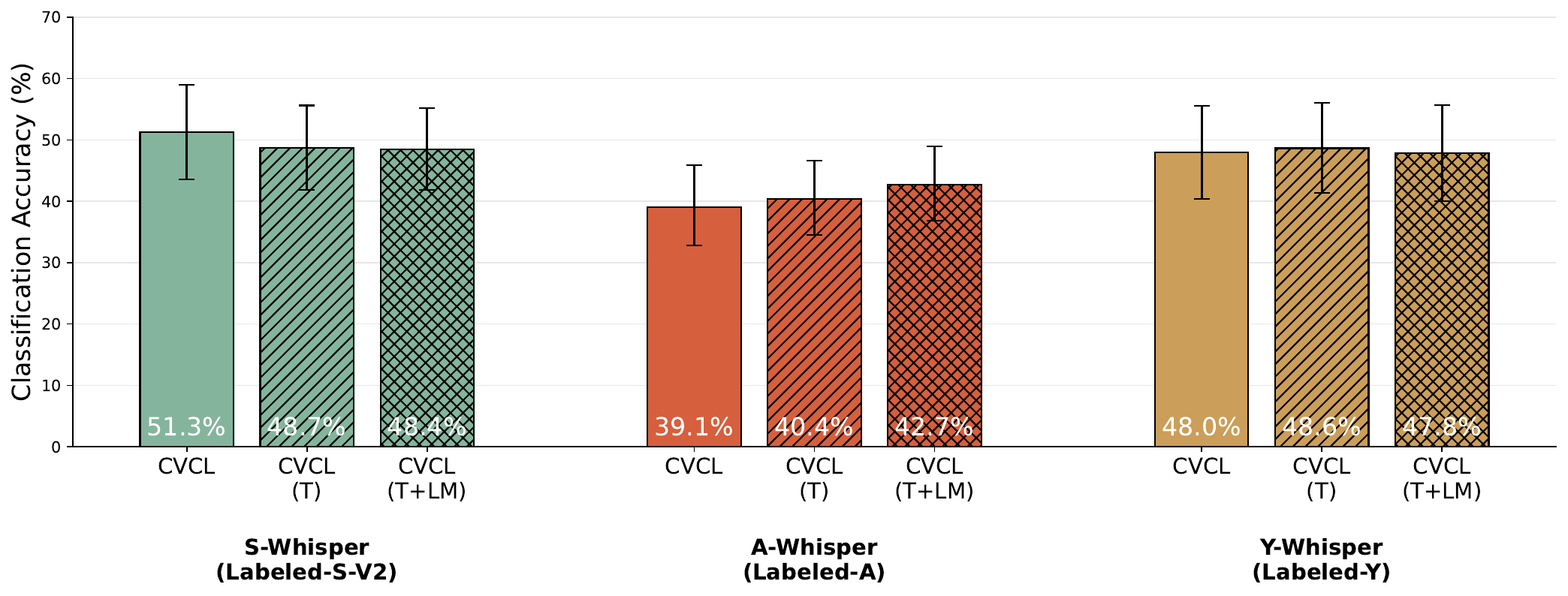}
  \caption{\textbf{Classification accuracy across models trained from Whisper transcriptions of each child in SAYCam.} Each model was evaluated on a new set containing 41 different categories, shared across the three evaluation datasets but using video frames derived from each child's videos (see Figure~\ref{fig:evaluation-frames} for some example evaluation frames). Models were trained with three random seeds. Error bars show bootstrapped 95\% confidence intervals over category-level accuracies.}
  \label{fig:manual_filtered_labeled_saycam_accuracy}
\end{figure}

Results for these evaluations are shown in Figure~\ref{fig:manual_filtered_labeled_saycam_accuracy}. At the highest level, we find that all model configurations and dataset splits, when evaluated on their corresponding evaluation split, achieve a classification accuracy higher than random chance (25\%). Overall, these results show that word-referent mappings are robustly learnable with the use of multimodal neural networks, not just in the case of one child's dataset \citep{vong2024grounded}, but instead across multiple independent children's datasets and across a variety of different model configurations. Looking more closely at the performance across dataset splits, we still see some qualitative differences in performance. Models trained on data from baby S and baby Y achieve higher levels of accuracy (at 51\% and 49\% respectively), while the best model configuration for baby A only achieves 43\% accuracy. Furthermore, and consistent with the previous set of results in comparing model variants, here we also find that no specific model configuration achieves superior classification performance. 

What drives the variation in classification performance across models trained on different children's datasets? A qualitative inspection of baby A's videos revealed that a large number of training examples contained parental discussions that were unrelated to the current visual context, providing noisy and irrelevant signals for grounded multimodal learning. To quantify the degree of vision-language alignment present in each child's training dataset, we used CLIP ViT-B/32 to compute the distribution of cosine similarities between training frames and their paired utterances. While mean CLIP similarities were nearly identical across children (baby S: 0.222, baby A: 0.220, baby Y: 0.216), the count of high-quality aligned examples (CLIP similarity $> 0.24$) varied substantially. Baby S's training data contained 9,320 highly-aligned pairs, baby A's training data had 3,223 pairs, and baby Y's training data had 3,514 pairs, corresponding to 7.60\%, 4.43\% and 4.94\% of their respective training datasets. The relative ranking of the number and proportion of high-quality training examples matches our observed classification performance, suggesting that the absolute number of high-quality examples, rather than the average quality, is the key driver behind successful multimodal word learning.

\subsection{Finding 6: Learned word-referent mappings also transfer across children and to novel image domains}

Finally, we examine generalization of word-referent mappings more broadly. First, we examine transfer to other children's visual environments (cross-child generalization), and second, to an entirely different image domain (out-of-distribution generalization).

\begin{figure}[t]
  \centering
\includegraphics[width=0.95\textwidth]{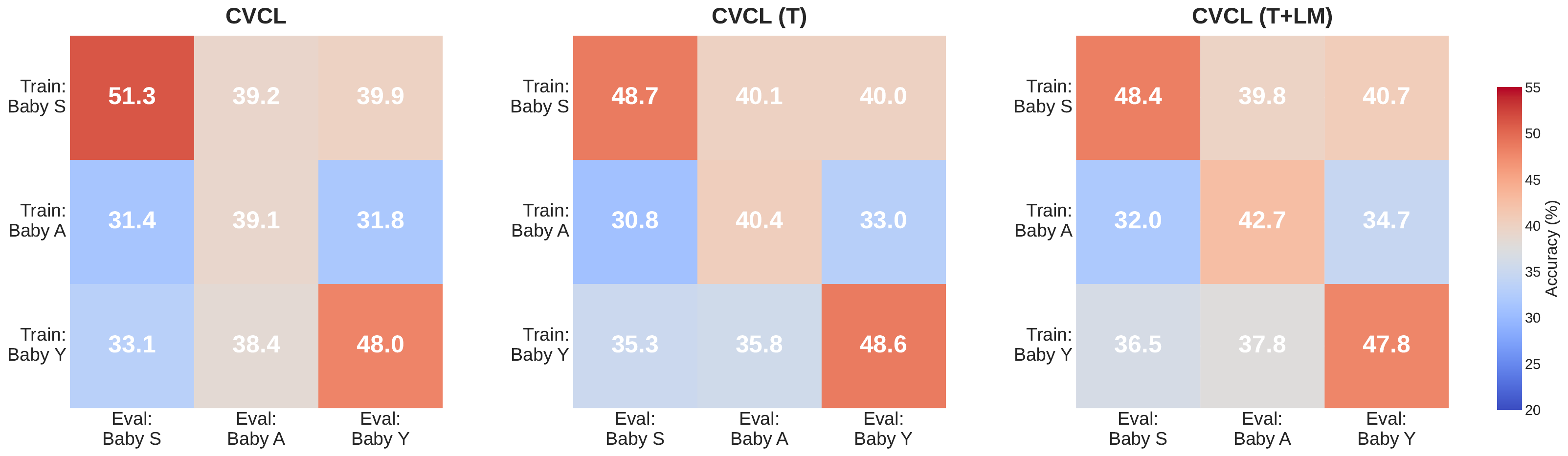}
  \caption{\textbf{Cross-child classification accuracy across models}. This figure presents classification accuracy scores when models trained on one child's data are evaluated on another child's evaluation set. Rows indicate the training dataset, while columns indicate the evaluation dataset. Diagonal entries represent the degree of within-child generalization, while off-diagonal entries represent the degree of cross-child generalization. All models show some degree of cross-child generalization, although some more than others.}
  \label{fig:cross-child-heatmaps}
\end{figure}

To examine whether models can generalize across children's visual environments, we tested models on each of the three evaluation datasets from the previous section, regardless of the dataset each model was trained on. That is, models trained on baby S's videos were evaluated on Labeled-A and Labeled-Y, et cetera. Results are shown in Figure~\ref{fig:cross-child-heatmaps}. The values along the diagonals represent the classification accuracy results from the previous section, corresponding to models trained and evaluated on the same child. However, the values on the off-diagonals represent the classification accuracy of cases where models were trained and evaluated on different children's data. Examining these results, we see a broadly similar pattern, while cross-child generalization shows reduced performance compared to within-child generalization. However, the resulting averaged classification accuracy scores remained well above the 25\% chance level, indicating that these learned word-referent mappings do indeed transfer over to other children's visual environments to some degree. Furthermore, and consistent with the results from the previous section, we also see better cross-child generalization performance for models trained on babies S and Y, compared to models trained on baby A.

\begin{figure}[t]
  \centering
\includegraphics[width=0.95\textwidth]{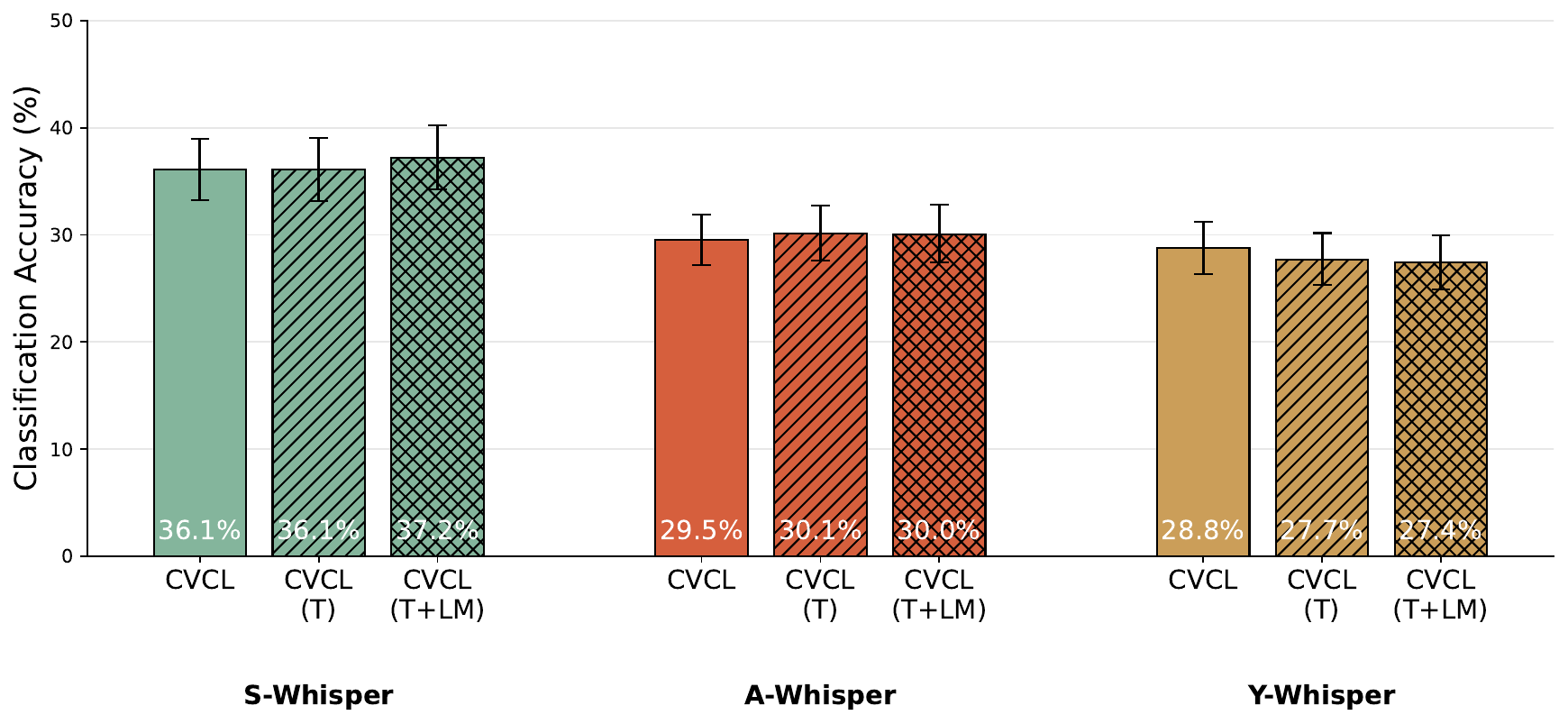}
\caption{\textbf{Out-of-distribution generalization to the Konkle Objects dataset}. Each model was evaluated on the same set of evaluation trials containing novel images on white backgrounds, across 60 different categories. Models trained on baby S show partial out-of-distribution performance, while models trained on babies A and Y are modestly above chance. Error bars show bootstrapped 95\% confidence intervals over category-level accuracies.}
  \label{fig:konkle-objects}
\end{figure}

In an even stricter test of out-of-distribution generalization, we also tested models on an adapted version of the same Konkle Objects dataset used in \cite{vong2024grounded}. All models were tested on the same set of evaluation trials, consisting of 60 categories (present in all three children's vocabularies) of naturalistic objects on white backgrounds across 1,784 trials, none of which were seen during training. Results are shown in Figure~\ref{fig:konkle-objects}. While models trained on data from baby S showed some limited generalization around 36 - 37\% classification accuracy, models trained on A and Y showed more limited generalization at 30\% and 28\% average classification accuracy respectively, modestly above chance-level. Consistent with our other evaluations, model architecture had minimal impact on performance, with training data characteristics remaining the primary factor driving performance differences. Together, these results show that all models can generalize beyond their training data, but that performance also progressively decreases from within-child to cross-child to out-of-distribution generalization, reflecting the increasing domain shift at each step. 

\section{Discussion}

In this work, we explored the robustness of simulated word learning from developmentally plausible datasets using multimodal neural networks with minimal inductive biases. We used automatic speech transcription methods to transcribe the entire SAYCam dataset, increasing the number of available transcribed utterances by a factor of 7, and use these automatically transcribed utterances to generate a number of multimodal training and evaluation datasets. We tested 3 different model configurations varying the type of language encoder and loss functions, against 5 different dataset splits across 3 babies. Overall, our results showed that all model configurations, across all three datasets, demonstrated that word-referent mappings were learnable using multimodal neural networks. Our findings build upon the results from \cite{vong2024grounded} --- although here we used a broader set of evaluation datasets, some with twice as many concepts --- highlighting that the kind and quantity of visual and linguistic information present in children's environments is indeed sufficient to acquire and generalize many word-referent mappings, without the need for additional linguistic inductive biases. Across multiple dataset splits, we show that the learned word embeddings generalize to visual instances across videos, across children and across image domains. Models successfully classify held-out frames from the same child, transfer to the visual environments of other children, and show modest above-chance performance on novel, naturalistic object stimuli, demonstrating some degree of transfer throughout.

While our results showed that the specific model configuration used for each dataset resulted in minor variations in classification performance, we found that no specific model configuration was dominant across all datasets. However, in the second set of evaluations, comparing performance across models trained across the three children with the same set of evaluation categories (although with different frames), we observed that performance on models trained on baby A were much worse compared to babies S or Y. One hypothesis we arrived at after watching a number of the videos is that there are far fewer instances of visual-linguistic alignment present in the training set. In particular, baby A's recordings (and transcribed utterances) often involved both parents having a discussion that was unrelated to the captured visual input, providing limited instances for multi-modal alignment for the model. While prior research has shown that children can acquire new words from overhearing, they often still required the novel words to be relevant in the current visual context \citep{Akhtar2005TheRO, Gampe2012EighteenmontholdsLN}.

While the use of automatic transcriptions for multimodal modeling has previously been explored \citep{roy2002learning,harwath2018jointly}, it has seen less uptake in the field of language acquisition, likely due to some of the difficulties in obtaining accurate and representative transcripts (although see \cite{long2024babyview} for another recent example). In this work, we find that training multimodal models from automatic transcripts enabled the training of separate models for each of the three children's dataset from the SAYCam corpora, that were previously inaccessible due to the efforts required for manual speech transcription. Furthermore, by comparing models trained from the same subset of videos from baby S either from manual vs. automated transcriptions, we find only a minor gap in classification performance. However, one advantage in our particular case for examining word-referent mappings is that the visual input of children changes relatively slowly given the egocentric nature of these datasets. Another advantage is that our evaluations focus on common object categories with sufficient exemplars during training, masking some of the issues with errors during transcription, or dealing with more uncommon words. Thus, Whisper's imperfect utterance timings, combined with the wide range from which we sample paired video frames, turns out to suffice for word learning. Finally, we note that our approach and current findings are specific to English, and generalizing this approach to other languages may prove to be challenging, given the high variability in accuracy of automated speech transcriptions across different languages, especially low-resource ones \citep{radford2023robust}.

Overall, our results provide a roadmap for studying multimodal language acquisition in developing children. Compared to \cite{vong2024grounded}, we explore datasets from multiple children and numerous model types and configurations, showing robustness to the acquisition of word-referent mappings across all of these configurations. Nevertheless, there still remain many other interesting research opportunities at the intersection of developmentally rich datasets and modern machine learning methods to further close the data-efficiency gap and understand how humans effectively learn language from so little. In particular, we hope future work will build on our learnability results to develop more psychologically and developmentally plausible cognitive models \citep{portelance2024roles}, incorporating more realistic attention and memory constraints, mechanisms for learning from actions and social interaction, and additional endogenous sources of linguistic input \citep{zhao2025missing}. Another natural extension is learning from raw speech rather than text transcriptions, which would allow models to leverage phonological and prosodic cues. While modeling from speech introduces additional challenges due to acoustic variability \citep{merkx2019language}, recent progress in speech-image learning suggests this is an increasingly tractable research direction \citep{harwath2016unsupervised, chrupala2022visually}.

\pagebreak
\subsubsection*{Acknowledgments}
We are grateful to the authors of the SAYCam article \citep{sullivan2022saycam}, and the volunteers who contributed to the data set. Thanks to Emin Orhan for training separate vision transformers from each child's data, which were used as the vision backbone for our multimodal neural networks. Additional thanks to Qiwen Zhang, Hailey He, Manli Zhao, Luyang Shang and Michael Picheny for their guidance on using Whisper and WhisperX to help automatically transcribe the SAYCam dataset, and their efforts to manually verify the quality of the transcriptions for this work. 

\bibliographystyle{apalike}
\bibliography{references}

\appendix
\section{Generating Evaluation Datasets}
\label{app:eval-datasets}

\subsection{Labeled-S-V2, Labeled-A and Labeled-Y}

In this work, we generated three new evaluation datasets (Labeled-S-V2, Labeled-A and Labeled-Y) to evaluate our multimodal models, which was performed using a mixture of both automatic and manual filtering. To accomplish this, we created new labeled datasets derived from the corresponding video frames from (primarily) the test set videos for each child,\footnote{The test set videos were unused during the training procedure, although for baby S there was some partial overlap.} creating three evaluation datasets consisting of 41 shared categories with 100 evaluation trials each, for a total of 4100 trials. 

These evaluation sets were created by first generating a set of target words or concepts to filter for. We generated this list by first considering the set of single word concepts from the MacArthur-Bates Communicative Development Inventory (MCDI), followed by filtering to (1) words whose concreteness score was higher than 4.8 (using concreteness scores from \cite{brysbaert2014concreteness}), (2) words present in all three children's vocabularies, and (3) words whose part-of-speech tag was either a \texttt{NOUN} or \texttt{VERB}, producing a set of 156 potential target labeled concepts to scan for.

Next, we used the largest available CLIP model (ViT-L/14@336px) \citep{radford2021learning}, and computed a similarity score (via cosine similarity) between every frame from each child's test set frames and the set of 156 target concepts (using the text embedding of the target label directly), providing us with an automated method to detect potential labeled frames. Using a similarity threshold of 0.24 (based on qualitative observations), we selected the highest scoring label for any given video frame above this threshold, which provided us with a set of labeled frames for each child. Using this as a starting point, we then performed a second round of manual filtering, by examining every automatically labeled instance to determine whether it was valid or not for the corresponding category label. After this manual filtering step, we performed a final filtering step to only include categories for which there were 10 or more instances in each of the three children's labeled datasets, leading to 41 categories in total.

Finally, we generated separate evaluation datasets for each child by sampling 100 trials per category with these corresponding labeled frames paired with three other randomly selected frames from other randomly selected categories from the same child, in a similar manner to the creation of the Labeled-S evaluation dataset, leading to three separate evaluation datasets consisting of 4100 trials each.

\subsection{Konkle Objects}

The Konkle Objects evaluation dataset was derived from a naturalistic image dataset containing multiple exemplars from 200 common object categories on white backgrounds from \cite{konkle2010conceptual}. Following \cite{vong2024grounded}, which used a 64-category subset from this dataset for evaluating models trained on baby S, we created an adapted version that was suitable for comparing across children to measure out-of-distribution generalization. Specifically, we selected the subset of object categories that were present in the vocabulary of all three children, resulting in 60 object categories. This choice of common categories ensured that models trained on any of the three children's datasets could be evaluated on this dataset.

For each of the exemplars in the 60 object categories, we generated 5 independent evaluation trials, consisting of one target image along with three additional foil images randomly selected from the other categories. This evaluation set consists of 1,784 evaluation trials across the 60 categories. 

\section{Generating the CLIP-filtered Validation Sets}
\label{app:clip-validation-sets}

Due to the limited signal from the noisy frame-utterance pairs that were automatically extracted, we needed a stronger signal to determine whether models were learning during the training process. To facilitate this, we leveraged CLIP to separately create filtered versions of the validation set for each data split, which was performed once as a pre-processing step prior to model training, ensuring that only frame-utterance pairs whose similarity exceeded a threshold were used to compute the validation loss. We used CLIP ViT-B/32 \citep{radford2021learning}, calculating the cosine similarity between a randomly selected frame and its associated utterance. We filtered the validation set to only include frame-utterance pairs whose similarity scores were greater than 0.24,\footnote{This threshold was determined via manual inspection from a small number of similarity scores.} thereby removing noisy or uninformative pairs for model validation during the training process. A comparison table for the size of the CLIP-filtered validation sets for each dataset split can be found in Table~\ref{tab:val-split-comparison}.

\section{Word Similarity Evaluation Results}
\label{app:word-similarity-evaluation}

In addition to our classification-based evaluations, we also conducted a separate set of word similarity evaluations to examine whether the use of a Transformer-based architecture, or language modeling objective produces richer semantic representations. Following the methodology of \cite{zhuang2024visual}, we used the word similarity benchmark MTest-3000 \citep{bruni2012distributional}, containing human similarity judgments for 3000 word pairs. We filtered this set down to only include pairs of words where both words had an age of acquisition (AoA) of 10 or below \citep{kuperman2012age}, and were present in the vocabulary of all three children, resulting in 725 word similarity pairs for evaluation.

\begin{figure}[h]
  \centering
  \includegraphics[width=0.95\textwidth]{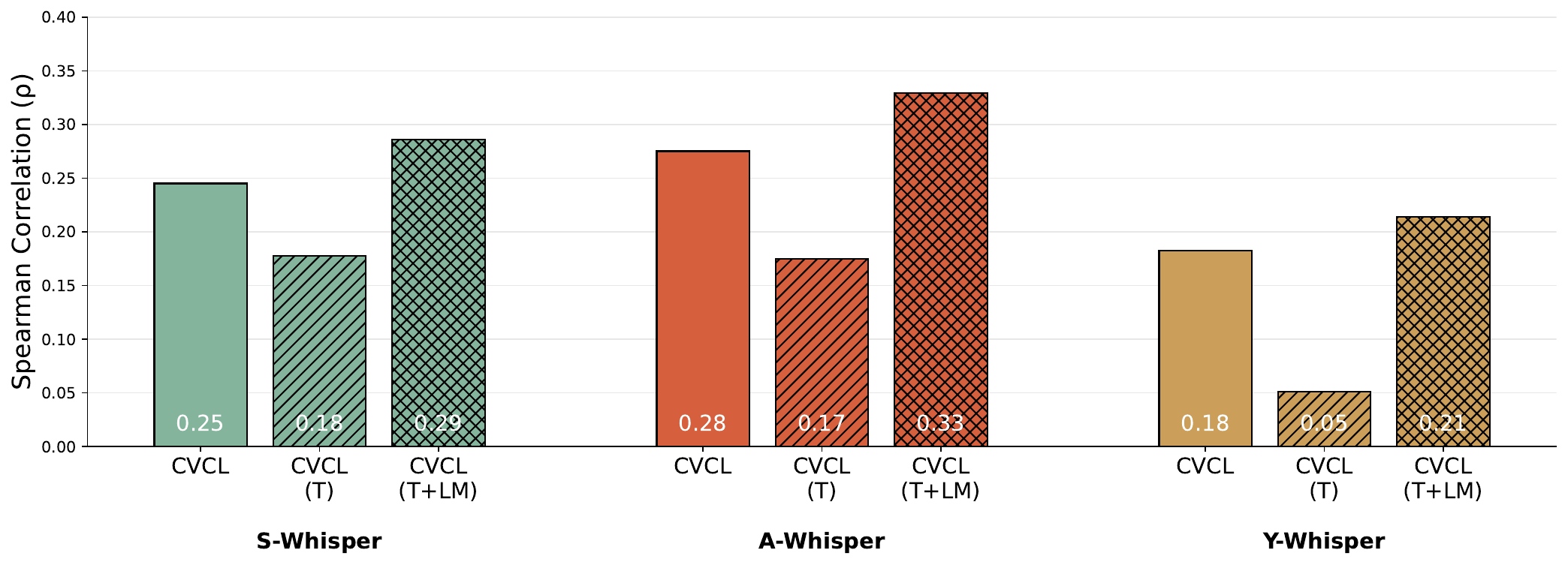}
\caption{\textbf{Word similarity evaluation across models and datasets.} Spearman correlation between model word embedding similarities and human similarity judgments, using the MTest-3000 word similarity benchmark filtered to word pairs with AoA $\leq$ 10 and present in each model's vocabulary. Joint models (CVCL+T+LM) consistently show higher correlations than contrastive-only models across all three children's datasets.}
 \label{fig:word-similarity}
\end{figure}

For each word pair, we computed the cosine similarity between the embeddings of the language encoder of each model, and measured the Spearman correlation between model similarities and human similarity judgments. Results are shown in Figure~\ref{fig:word-similarity}. Across each child, we see that the joint models (CVCL+T+LM) showed stronger correlations with human similarity judgments, compared to either of the contrastive-only models (CVCL and CVCL+T), indicating that the additional language modeling objective captures some additional semantic structure that contrastive-only models do not.

\section{Evaluation Dataset Categories}
\label{app:eval-categories}

\textbf{Labeled-S Categories} (22 categories): ball, basket, car, cat, chair, computer, crib, door, floor, foot, ground, hand, kitchen, paper, puzzle, road, room, sand, stairs, table, toy, window \\

\noindent \textbf{Labeled-S-V2, Labeled-A and Labeled-Y Categories} (41 categories): arm, ball, banana, bed, bedroom, blanket, book, boots, bottle, bowl, box, bread, bucket, cereal, cow, cup, door, finger, foot, hair, horse, jeans, kitchen, knee, leg, monkey, napkin, oven, paper, pen, pillow, porch, shirt, shoe, shoulder, sock, sun, table, towel, toy, window \\

\noindent \textbf{Konkle Objects Categories} (60 categories): airplane, apple, bagel, ball, balloon, basket, bed, bell, bike, bill, bird, boot, bottle, bowl, bucket, butterfly, button, cake, camera, cat, chair, cheese, clock, cookie, crib, dog, doll, fan, guitar, hat, jacket, juice, key, knife, leaves, lock, meat, necklace, pants, pen, phone, pipe, pizza, ring, rock, rug, sandwich, shoe, socks, spoon, stool, tape, tent, train, tree, trumpet, turtle, tv, umbrella, watch

\pagebreak
\section{Tables}
\label{app:tables}

\begin{table}[H]
    \begin{adjustbox}{center}
\small
\begin{tabular}{lcccc}
\toprule
\textbf{Dataset} & \textbf{Train} & \textbf{Validation} & \textbf{Test} & \textbf{Total} \\
\midrule
\multicolumn{5}{l}{\textbf{S-Whisper-2022}} \\
Number of videos & 348 & 21 & 17 & 386 \\
Number of utterances & 39649 & 2348 & 2045 & 44042 \\
Avg. utterance length & 4.76 & 4.97 & 4.70 & 4.77 \\
Number of extracted frames & 363782 & 22509 & 19347 & 405638 \\
Avg. frames per utterance & 9.18 & 9.59 & 9.46 & 9.21 \\
Total words & 188610 & 11674 & 9621 & 209905 \\
Vocabulary size & - & - & - & 2268 \\
\midrule
\multicolumn{5}{l}{\textbf{S-Whisper-Disjoint}} \\
Number of videos & 623 & 100 & 286 & 1009 \\
Number of utterances & 91953 & 10327 & 33715 & 135995 \\
Avg. utterance length & 4.57 & 4.84 & 4.74 & 4.64 \\
Number of extracted frames & 799615 & 98097 & 307541 & 1205253 \\
Avg. frames per utterance & 8.70 & 9.50 & 9.12 & 8.86 \\
Total words & 420609 & 50024 & 159881 & 630514 \\
Vocabulary size & - & - & - & 3375 \\
\midrule
\multicolumn{5}{l}{\textbf{S-Whisper}} \\
Number of videos & 916 & 46 & 47 & 1009 \\
Number of utterances & 122571 & 6888 & 6536 & 135995 \\
Avg. utterance length & 4.64 & 4.59 & 4.60 & 4.64 \\
Number of extracted frames & 1088036 & 60539 & 56678 & 1205253 \\
Avg. frames per utterance & 8.88 & 8.79 & 8.67 & 8.86 \\
Total words & 568852 & 31606 & 30056 & 630514 \\
Vocabulary size & - & - & - & 3856 \\
\midrule
\multicolumn{5}{l}{\textbf{A-Whisper}} \\
Number of videos & 299 & 38 & 44 & 381 \\
Number of utterances & 72810 & 9357 & 8810 & 90977 \\
Avg. utterance length & 4.65 & 5.10 & 4.65 & 4.69 \\
Number of extracted frames & 689794 & 90374 & 85755 & 865923 \\
Avg. frames per utterance & 9.47 & 9.66 & 9.73 & 9.52 \\
Total words & 338476 & 47698 & 40949 & 427123 \\
Vocabulary size & - & - & - & 4199 \\
\midrule
\multicolumn{5}{l}{\textbf{Y-Whisper}} \\
Number of videos & 246 & 34 & 31 & 311 \\
Number of utterances & 71101 & 8688 & 8816 & 88605 \\
Avg. utterance length & 4.84 & 4.55 & 4.85 & 4.81 \\
Number of extracted frames & 670820 & 82276 & 85271 & 838367 \\
Avg. frames per utterance & 9.43 & 9.47 & 9.67 & 9.46 \\
Total words & 344221 & 39511 & 42747 & 426479 \\
Vocabulary size & - & - & - & 4238 \\
\bottomrule
\end{tabular}
\end{adjustbox}
\caption{\textbf{Dataset Descriptives}}
\label{tab:dataset-descriptives}
\end{table}

\pagebreak
\begin{table}[ht]
  \centering
  \small
\begin{tabular}{lrrr}
\toprule
  \textbf{Dataset} & \textbf{Val} & \textbf{CLIP-Filtered Val} \\
  \midrule
  \multicolumn{3}{l}{\textbf{S-Whisper-2022}} \\
  Utterances & 2,348 & 350 \\
  Total Frames & 22,509 & 3,478 \\
  \midrule
  \multicolumn{3}{l}{\textbf{S-Whisper-Disjoint}} \\
  Utterances & 10,327 & 1,602 \\
  Total Frames & 98,097 & 16,482 \\
  \midrule
  \multicolumn{3}{l}{\textbf{S-Whisper}} \\
  Utterances & 6,888 & 932 \\
  Total Frames & 60,539 & 8,899 \\
  \midrule
  \multicolumn{3}{l}{\textbf{A-Whisper}} \\
  Utterances & 9,357 & 925 \\
  Total Frames & 90,374 & 9,494 \\
  \midrule
  \multicolumn{3}{l}{\textbf{Y-Whisper}} \\
  Utterances & 8,688 & 870 \\
  Total Frames & 82,276 & 9,142 \\
  \bottomrule
\end{tabular}%
\caption{\textbf{Comparison of Standard vs. CLIP-Filtered Validation Splits}}
\label{tab:val-split-comparison}
\end{table}

\pagebreak
\begin{table}[H]
\begin{adjustbox}{center}
\begin{tabular}{lcccc}
\toprule
\multirow{2}{*}[2.5ex]{\textbf{Model}} & \shortstack{\textbf{CVCL}\\ Vong et al. (2024)} & \shortstack{\textbf{CVCL}\\\textbf{ }} & \shortstack{\textbf{CVCL}\\ \textbf{(T)}} & \shortstack{\textbf{CVCL}\\ \textbf{(T+LM)}} \\
\midrule
\multicolumn{5}{l}{\textit{Architecture}} \\
Vision encoder & ResNeXt-50 & ViT-B/16 & ViT-B/16 & ViT-B/16 \\
Embedding size & 512 & 512 & 512 & 512 \\
Language encoder & Embedding & Embedding & Transformer Decoder & Transformer Decoder \\
Number of attention heads & - & - & 8 & 8 \\
Number of layers & - & - & 2 & 2 \\
Positional encoding & - & Learned, absolute & Learned, absolute & Learned, absolute \\
Max sequence length & 25 & 48 & 48 & 48 \\
\multirow{2}{*}[2.5ex]{Frame sampling method} & \shortstack{Sample\\ single frame} & \shortstack{Sample\\ single frame} & \shortstack{Sample\\ single frame} & \shortstack{Sample\\ single frame} \\
Pre-trained vision encoder & True & True & True & True \\
Fine-tune vision encoder & False & False & False & False \\
Image data augmentation & True & True & True & True \\
\midrule
\multicolumn{5}{l}{\textit{Loss Configuration}} \\
\multirow{2}{*}[2.5ex]{Loss} & \multirow{2}{*}[2.5ex]{Contrastive} & \multirow{2}{*}[2.5ex]{Contrastive} & \multirow{2}{*}[2.5ex]{Contrastive} & \shortstack{Contrastive and\\Language Modeling} \\
Fixed temperature & True & True & True & True \\
Initial temperature & 0.07 & 0.07 & 0.07 & 0.07 \\
$\lambda_\text{Contrastive}$ & - & - & - & 0.3 \\
\midrule
\multicolumn{5}{l}{\textit{Training Configuration}} \\
Train batch size & 8 & 16 & 64 & 64 \\
Validation batch size & 8 & 8 & 8 & 8 \\
CLIP-filtered Validation & False & True & True & True \\
Optimizer & AdamW & AdamW & AdamW & AdamW \\
Learning rate & 1e-4 & 1e-4 & 1e-5 & 1e-5 \\
\multirow{2}{*}[2.5ex]{LR Scheduler} & \multirow{2}{*}[2.5ex]{ReduceLROnPlateau} & \shortstack{Linear Warmup with\\Cosine Annealing} & \shortstack{Linear Warmup with\\Cosine Annealing} & \shortstack{Linear Warmup with\\Cosine Annealing} \\
Warmup steps & - & 5000 & 5000 & 5000 \\
Weight decay & 0.1 & 0.1 & 0.1 & 0.1 \\
Vision dropout & - & 0.1 & 0.1 & 0.1 \\
Vision layer norm & - & True & True & True \\
Language dropout & - & 0.1 & 0.1 & 0.1 \\
Num epochs & 400 & 100 & 100 & 100 \\
Early Stopping & Validation loss & Validation loss & Validation loss & Validation loss \\
\bottomrule
\end{tabular}
\end{adjustbox}
\caption{\textbf{Comparison of model configurations}}
\label{tab:model-configurations}
\end{table}

\begin{table}[ht]
  \centering
  \small
\begin{tabular}{lcccccc}
\toprule
\textbf{Evaluation Category} & \multicolumn{2}{c}{\textbf{Labeled-S-V2}} & \multicolumn{2}{c}{\textbf{Labeled-A}} & \multicolumn{2}{c}{\textbf{Labeled-Y}} \\
& \textbf{Frames} & \textbf{Videos} & \textbf{Frames} & \textbf{Videos} & \textbf{Frames} & \textbf{Videos} \\
\midrule
Arm & 91 & 42 & 89 & 20 & 67 & 19 \\
Ball & 92 & 23 & 36 & 7 & 87 & 11 \\
Banana & 71 & 9 & 52 & 4 & 75 & 5 \\
Bed & 53 & 18 & 30 & 9 & 43 & 5 \\
Bedroom & 74 & 13 & 17 & 6 & 49 & 4 \\
Blanket & 60 & 10 & 26 & 5 & 37 & 11 \\
Book & 83 & 16 & 89 & 9 & 40 & 8 \\
Boots & 32 & 4 & 15 & 4 & 69 & 11 \\
Bottle & 86 & 19 & 69 & 9 & 81 & 15 \\
Bowl & 64 & 10 & 51 & 6 & 87 & 13 \\
Box & 90 & 16 & 75 & 9 & 44 & 5 \\
Bread & 34 & 12 & 12 & 1 & 48 & 6 \\
Bucket & 51 & 9 & 13 & 6 & 10 & 4 \\
Cereal & 15 & 4 & 44 & 3 & 77 & 8 \\
Cow & 51 & 5 & 14 & 6 & 24 & 1 \\
Cup & 65 & 10 & 79 & 8 & 95 & 15 \\
Door & 74 & 26 & 33 & 12 & 60 & 11 \\
Finger & 83 & 46 & 71 & 19 & 57 & 18 \\
Foot & 95 & 31 & 79 & 20 & 78 & 18 \\
Hair & 30 & 16 & 16 & 6 & 36 & 6 \\
Horse & 36 & 4 & 30 & 3 & 12 & 5 \\
Jeans & 71 & 22 & 56 & 11 & 31 & 8 \\
Kitchen & 95 & 29 & 30 & 8 & 97 & 22 \\
Knee & 32 & 11 & 11 & 6 & 10 & 8 \\
Leg & 87 & 16 & 83 & 11 & 77 & 9 \\
Monkey & 60 & 6 & 59 & 3 & 15 & 2 \\
Napkin & 29 & 7 & 18 & 3 & 26 & 5 \\
Oven & 35 & 7 & 11 & 2 & 37 & 8 \\
Paper & 40 & 12 & 19 & 4 & 44 & 5 \\
Pen & 10 & 1 & 45 & 4 & 22 & 2 \\
Pillow & 17 & 1 & 42 & 6 & 48 & 5 \\
Porch & 92 & 25 & 56 & 7 & 85 & 3 \\
Shirt & 68 & 13 & 21 & 2 & 14 & 9 \\
Shoe & 94 & 15 & 70 & 8 & 75 & 8 \\
Shoulder & 66 & 21 & 66 & 9 & 21 & 5 \\
Sock & 65 & 13 & 52 & 8 & 70 & 15 \\
Sun & 54 & 14 & 29 & 10 & 20 & 3 \\
Table & 83 & 21 & 22 & 8 & 93 & 18 \\
Towel & 18 & 9 & 12 & 3 & 10 & 5 \\
Toy & 92 & 27 & 91 & 12 & 98 & 20 \\
Window & 81 & 23 & 87 & 10 & 66 & 11 \\
\bottomrule
\end{tabular}
\caption{\textbf{Number of unique exemplars and unique videos per evaluation category in Labeled-S-V2, Labeled-A and Labeled-Y Evaluation Datasets}}
\label{tab:unique-exemplars}
\end{table}

\end{document}